\pdfoutput=1
\documentclass[11pt]{article}

\usepackage{EMNLP2022}
\usepackage{times}
\usepackage{latexsym}

\usepackage{fancyhdr}
\fancypagestyle{alim}{\fancyhf{}\fancyfoot[C]{\emph{\small Published as a conference paper at EMNLP 2022 (Findings)}}}
\usepackage[T1]{fontenc}
\usepackage[utf8]{inputenc}

\usepackage{microtype}

\usepackage{inconsolata}

\usepackage{amsmath}
\usepackage{amsfonts}

\usepackage{booktabs}
\usepackage{makecell} % two lines in one table cell
\usepackage{multirow}
\usepackage{multicol}
\usepackage{algorithm}
\usepackage{adjustbox}
\usepackage{float}
\usepackage{caption}
\usepackage{subcaption}
\usepackage{graphics}
\newsavebox{\tempbox}
\usepackage{amsmath,amsthm,amssymb}
\usepackage{mathtools}
\usepackage{xspace}
\usepackage{algorithm}
\usepackage{algpseudocode}

\title{Continuation KD: Improved Knowledge Distillation through the Lens of Continuation Optimization}

\author{Aref Jafari$^{1}$ \quad Ivan Kobyzev$^{2}$ \quad Mehdi Rezagholizadeh$^{2}$ \quad \\ \textbf{Pascal Poupart$^1$} \quad \textbf{Ali Ghodsi$^{1}$} \\
  $^1$University of Waterloo \\
  $^2$Huawei Noah's Ark Lab \\
  \small \texttt{\{aref.jafari, ppoupart, ali.ghodsi\}@uwaterloo.ca}\\
  \small \texttt{\{ivan.kobyzev, mehdi.rezagholizadeh\}@huawei.com} 
}

\begin{document}
\maketitle
\pagestyle{fancy}
\fancyhf{}
\thispagestyle{alim}
% \pagestyle{fancy}
% \fancyhf{}
% % \fancyhfoffset[L]{1cm} % left extra length
% % \fancyhfoffset[R]{1cm} % right extra length
% \cfoot{\ifnum\thepage=1 Published as a conference paper at EMNLP 2022 (Findings)\else \fi}
% \cfoot{Published as a conference paper at EMNLP 2022 (Findings)}

\begin{abstract}
% Compressing neural network models is an important problem in machine learning. Small models can empower edge devices to expand their services among users with low computational resources.
Knowledge Distillation (KD) has been extensively used for natural language understanding (NLU) tasks to improve a small model's (a student) generalization by transferring the knowledge from a larger model (a teacher). Although KD methods achieve state-of-the-art performance in numerous settings, they suffer from several problems limiting their performance. It is shown in the literature that the capacity gap between the teacher and the student networks can make KD ineffective. Additionally, existing KD techniques do not mitigate the noise in the teacher's output: modeling the noisy behaviour of the teacher can distract the student from learning more useful features. We propose a new KD method that addresses these problems and facilitates the  training compared to previous techniques. Inspired by continuation optimization, we design a training procedure that optimizes the highly non-convex KD objective by starting with the smoothed version of this objective and making it more complex as the training proceeds. Our method (Continuation-KD) achieves state-of-the-art performance across various compact architectures on NLU (GLUE benchmark) and computer vision tasks (CIFAR-10 and CIFAR-100).
\end{abstract}

\section{Introduction}

Deep neural networks have achieved great success in many challenging tasks including the ones in natural language processing ~\cite{vaswani2017attention, NEURIPS2020_1457c0d6} and computer vision ~\cite{dosovitskiy2020image}. 
Most successful neural networks are usually large and overparametrised~\cite{liu2019roberta, devlin2018bert}. The big size of these models prevents them from being deployed on computers with low computational power such as edge devices. Compressing the models can empower us to provide a variety of machine learning-based services offline on low-resource machines. This issue is even more severe in models designed for NLP. Although, after the introduction of the Transformers ~\cite{vaswani2017attention} the performance of models in this field improved dramatically, the size of the models increased exponentially. Nowadays, some of these models have more than 100 billion parameters ~\cite{brown2020language}, and they are still increasing. 

One way to address the expensive computational complexity of deep
networks and their over-parameterization is neural model compression ~\cite{jacob2018quantization, tjandra2018tensor,bie2019simplified}. Among all of the compression methods, Knowledge Distillation (KD) ~\cite{hinton2015distilling} is one of the prominent techniques which have been used to compress a variety of models in different deep learning applications such as Natural Language Processing ~\cite{clark-etal-2019-bam, sunmobilebert, jiao2019tinybert,rashid2020towards, rashid2021mate,kamalloo2021not,haidar2021rail,wu2021universal}, speech processing ~\cite{yun2020regularizing, chebotar2016distilling}, and computer vision ~\cite{mirzadeh2019improved, guo2020online}.
In KD, we have a large accurate network, which is called a teacher, and a small network, which is called a student, that we desire to train. The innate capacity gap between the student and the teacher was speculated~\cite{lopez2015unifying} to impede the training and was addressed by multiple works \cite{mirzadeh2019improved, jafari2021annealing}. However, the existing solutions have several complications: for example, \cite{mirzadeh2019improved} requires an extra intermediate network to be trained, and \cite{jafari2021annealing} has a rigid two-stage structure that requires the careful design for successful training. Moreover, none of these techniques  is robust against the noisy
data and noisy teacher’s outputs, which can
distract the student with the limited capacity from learning more useful features and make it overfit to the noise. 

We propose a new solution, Continuation KD, inspired by the continuation method from optimization and nonlinear equations. During the training, we gradually move from the smoothed objective function, which is robust to overfitting to the noise, to the original highly non-convex function. We conduct extensive experiments on the GLUE benchmark for DistilRoBERTA (6-layer)~\citep{sanh2019distilbert} and BERT-small (4-layers) \cite{turc2019wellread} core models and show a significant improvement over the previous baselines. Besides that, we demonstrate that in the computer vision setting continuation KD also outperforms its competitors. 
Overall, our contributions are the following:
\begin{enumerate}
    \item We proposed a novel KD technique based on the Continuation method, which gradually increases the complexity of the loss function and provides a better optimization for all knowledge distillation scenarios in both computer vision and NLP. 
    
    \item We implement a loss function like hinge loss function which makes a student trained with our technique robust against the teacher's output noise.  
    
    \item Our proposed method is simple and, unlike its competitors, does not have several stages. This feature makes our method stable and efficient. 
\end{enumerate}

\section{Related Works}
\subsection{Knowledge Distillation (KD)}
Knowledge Distillation~\cite{hinton2015distilling} is a well-known neural model compression method.
Despite the success of the original method, it has been shown in the literature ~\cite{lopez2015unifying, mirzadeh2019improved} that the large gap between the size of the student and the teacher networks makes KD ineffective. 
To address this \textit{capacity gap} problem, \citet{mirzadeh2019improved} proposed the teacher assistant knowledge distillation (TAKD) method.However, this technique is computationally expensive as it requires training multiple TA networks for a task. Moreover, the errors of the TAs can accumulate and transfer to the student.
To alleviate these problems, ~\citet{jafari2021annealing} proposed Annealing-KD that achieved state-of-the-art performance on NLU and computer vision tasks. Although this method could handle the capacity gap problem, it is still not robust against the noisy data and noisy teacher's outputs. Also, two phases of the training require some \textit{a priori} decisions on when to switch from the first phase to the second one.

\subsection{Continuation Optimization}
Continuation method was first proposed as a numerical method for solving nonlinear equations~\cite{continuation_53} and then it was adopted as a heuristic for nonconvex optimization~\cite{continuation_00}. 
The main intuition of the continuation method is to include the problem we are trying to solve in a continuous family of problems, such that one of the members of this family is easy to solve and its solution could be pulled over to give an approximate solution of the original problem. 

Machine Learning community applied the continuation idea for training neural networks. \citet{mobahi_fisher_2015} proposed the first theoretical analysis of the bound on the approximate solution given by the continuation optimization. 
\citet{mollifying} suggested optimizing highly non-convex neural networks
by starting with a smoothed objective function and making it more complex over the training. We adapted this general method for Knowledge Distillation.

\section{Background}
\paragraph{Vanilla-KD}
The original knowledge distillation ~\cite{hinton2015distilling}  trains a small network (a student) by using two guiding signals: the hard labels coming from the training dataset, and the predictions of a large network pre-trained on the same task (a teacher) which is known as soft labels. To achieve this goal, Vanilla-KD utilizes a particular loss function which is a linear combination of two losses. The first one is a cross-entropy between the softmax output of the student and hard labels, and the second one is a KL-divergence between the softened version of the softmax outputs of the student and the teacher networks.
Equation \ref{eq:1} explains this loss function in details:
\begin{equation} \label{eq:1}
\begin{split}
    \mathcal{L}_{KD} = \lambda CE(y, \sigma\big(z_S(x)\big) + \\ 
    (1-\lambda) KL(\sigma(\frac{z_T(x)}{\mathcal{\tau}}), \sigma(\frac{z_S(x)}{\mathcal{\tau}}))
    \end{split}
\end{equation}
Here $CE(.)$ is the cross-entropy function, $KL(.)$ is the KL-divergence function, $z_T(.)$ and $z_S(.)$ are the teacher and student logits, $\sigma(.)$ is the softmax function, and $\mathcal{\tau}$ is the softening parameter. Also, $\lambda$ is a hyper-parameter between $[0, 1]$ which indicates the amount of contribution of each loss function. Minimizing the above loss function decreases the distance between both underlying function and the teacher model. In Vanilla-KD usually we assume that the teacher is a good approximation of the underlying function. 

\paragraph{TAKD}
In TAKD method~\cite{mirzadeh2019improved} the teacher model first trains an intermediate model with a slightly smaller capacity, which is called teacher assistant (TA), by utilizing Vanilla-KD. Then the TA model trains the student model with a small capacity by using Vanilla-KD again. TAKD tries to fill the capacity gap between the teacher and the student models by introducing the TA model but this gap can be still large. As mentioned in ~\cite{mirzadeh2019improved}, a better idea is to use hierarchical TAs to have a smoother knowledge transfer from the teacher to the student. 

\paragraph{Annealing-KD}
For controlling the complexity of the teacher model, instead of using multiple TA networks, Annealing-KD~\citep{jafari2021annealing} adds an annealed dynamic temperature factor to the output of the teacher. By using this factor, Annealing-KD reduces the sharpness of the teacher at the beginning of the training process. Then it increases the sharpness of the teacher gradually during the training. Therefore, since the complexity of the teacher increases gradually during the training time, the teacher knowledge transfers much more smoothly to the student than in TAKD.

Annealing-KD has two stages of training where in the first stage, a student learns from a teacher for $k$ epochs. During this stage, Annealing-KD matches the student's logits to the teacher's logits by using a mean square error loss function. At the beginning of the training the temperature factor sets to a high value to apply the maximum smoothing to the output of the teacher and then it decreases gradually until there is no smoothing effect remains on the output of the teacher.
Formally, one first defines a monotonically increasing function $\phi: \mathbb{N} \to [0,1]$, going from zero at the beginning of the training to one at the end.  Then, the student  loss for stage 1 is:
 \begin{equation}
\label{eq:akd}
    \mathcal{L}(i) = || z_S(x) - \phi(i)z_T(x) ||_2^2,
\end{equation}
where $i$ is the training epoch, $z_S(x)$ and $z_T(x)$ are logit outputs of the student and the teacher respectively. 
In the second stage, Annealing-KD gets the best checkpoint of the first stage and finetune it on the hard labels from the dataset by using a cross-entropy loss for $m$ epochs. The hyperparameters $k$ and $m$ must be chosen before the training.

\begin{figure*}[bt]
\label{figur1:1}
 \centering
  \includegraphics[width=0.8\textwidth]{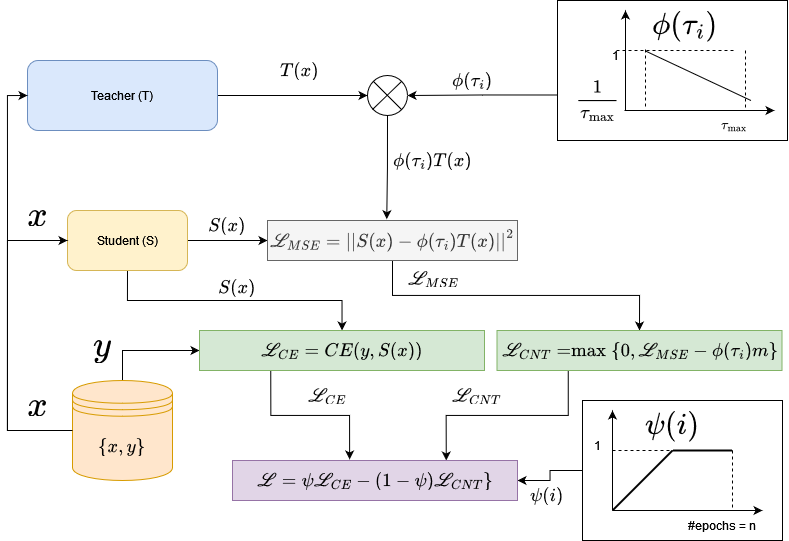}
 
  \caption{Principle diagram illustrating different components of Continuation-KD. The main loss function (purple box) is a composition of two losses - cross entropy loss $\mathcal{L}_{CE}$  and continuation loss $\mathcal{L}_{CNT}$ (green boxes). Because of the smoothing in $\mathcal{L}_{CNT}$ with the dynamic factor $\phi$, it is an easier objective to optimize than $\mathcal{L}_{CE}$. During the training, Continuation KD gradually moves from the easier objective to more complex objective with aid of the dynamic factor $\psi$.
}
  \label{fig1}
\end{figure*}

\section{Methodology}
In this section, we describe our Continuation-KD technique, which addresses both the capacity gap problem and the lack of robustness against the noise in the teacher's output. 
Continuation-KD uses a loss function with two objectives (Eq.~\ref{eq:Continuation}). The first objective $\mathcal{L}_{CE}$ is a cross-entropy loss term that trains the student based on the given hard labels.
The second objective $\mathcal{L}^{CNT}_{KD}$ is our proposed annealed hinge loss function that  gradually trains the student to mimic the behaviour of the teacher. 
Inspired by the continuation method \cite{mollifying}, Continuation-KD starts with an easy objective to train at the beginning of training. As the training proceeds, the whole objective function becomes more and more complex. Formally, the loss function of the Continuation-KD is defined as follows: 
\begin{equation} \label{eq:Continuation}
\begin{split}
    \mathcal{L} = (\psi(i))\mathcal{L}_{CE} +  
    (1-\psi(i))\mathcal{L}^{CNT}_{KD} 
    \end{split}
\end{equation}
where $1 \leq i \leq n$ indicates the epoch index with the maximum number of epochs $n$. The $0 \leq \psi(i) \leq 1$ is an increasing function between 0 and 1 where $\psi(1)=0$ at the beginning and it increases during the training. $L^{CNT}_{KD}$ defines as:
\begin{equation} \label{eq:Continuation_loss}
\begin{split}
    \mathcal{L}^{CNT}_{KD} =  \max\{0, \lVert z_S - \phi(\mathcal{T}_i)  z_T\rVert^{2}_{2} - m \text{ } \phi(\mathcal{T}_i)\}
    \end{split}
\end{equation}
    where $z_S$ and $z_T$ are the output logits of the student and teacher networks, respectively; $m$ is the margin factor; $1 \leq \mathcal{T}_i \leq \mathcal{T}_{max}$ is the temperature factor,  $\mathcal{T}_{max}$ is the maximum temperature, and $0 \leq \phi(\mathcal{T}_i) \leq 1$ is an increasing function. We define this function as:

\begin{equation} \label{eq:Continuation_phi}
\begin{split}
    \phi(\mathcal{T}_i) = 1-\frac{\mathcal{T}_i-1}{\mathcal{T}_{\text{max}}}, 1 \leq \mathcal{T}_i \leq \mathcal{T}_{\text{max}}, \mathcal{T}_i\in \mathbb{N}. 
    \end{split}
\end{equation}

Note that in Eq.~\ref{eq:Continuation_loss}, $\lVert z_S - \phi(\mathcal{T}_i)  z_T\rVert^{2}_{2}$ is a mean square loss between the student's logits and the annealed version of the teacher's logits.  Also, $\max\{0, \lVert z_S - \phi(\mathcal{T}_i)  z_T\rVert^{2}_{2} - \phi(\mathcal{T}_i)m\}$ is a hinge loss with an annealed margin $\phi(\mathcal{T}_i)m$. This loss function avoids penalizing negligible differences (the ones less than $m \text{ }\phi(\mathcal{T}_i)$) between the outputs of the student and teacher. This feature helps the student to learn a meaningful behaviour of the teacher rather than focusing on higher frequency fluctuations.

At the beginning of training, we set $\mathcal{T}_1 = \mathcal{T}_{max}$ which leads to the most softened version of the teacher's output $(\phi(\mathcal{T}_1) = \frac{1}{\mathcal{T}_{max}})$. Since we have $\psi(1)=0$, the student only learns the behaviour of the teacher's smoothest version, which is an easy target to learn.
Then, during training, we decrease the temperature,  
At this phase, functions $\psi(i)$ and $\phi(\mathcal{T}_i)$ are both increasing which in turn leads to increasing the sharpness of the teacher and smoothly shifting from the hinge loss to the cross-entropy loss. Both of these operations increase the complexity of the whole loss function. Note that at the function $\phi(\mathcal{T}_i)$ also anneals margin $m$. Its reason is that smoothing the teacher with  $\phi(\mathcal{T}_i)$ damps its noise as well. Therefore we damp the margin $m$ with  $\phi(\mathcal{T}_i)$ to apply a margin proportional to the amount of noise in the smoothed version of the teacher. Figure \ref{fig1} visualize different components of continuation-KD. 

Also note that, if we set $m=0$ and $\psi(i)$ to the step function in Eq~\ref{eq:step_fnc}, then Continuation-KD becomes identical to Annealing-KD, where $k$ is the number of epochs in the first stage and $n-k$ is the number of epochs in the second stage. This fact shows that the Annealing-KD is actually a special case of our Continuation-KD. 

\begin{equation}
    \begin{split}
        &   \psi(i) =    \begin{cases}
    0,              &   1 \leq i \leq k\\
    1,              & k < i  \leq n
\end{cases}  
    \end{split}
    \label{eq:step_fnc}
\end{equation}

Algorithm \ref{alg} demonstrates the details of Continuation-KD. It requires a student $S$, a teacher $T$, a dataset $D$, max temperature $\mathcal{T}_{max}$, number of epochs $n$, an increasing function $\psi$, and margin $m$ as inputs and returns the trained student at the output. At the beginning, it sets variables $\mathcal{T} = \mathcal{T}_{max}$
to get the maximum smoothness of the teacher. Also, variable $k$ indicates the number of epochs before updating $\mathcal{T}$during training. $\Phi$ and $\Psi$ are the output values of $\phi(i)$ and $\psi(i)$. Function $\Call{Get-Mini-Batch}{D}$ retrieves a mini-batch $(X, Y)$ from the dataset $D$. Then these data samples feed into loss functions in the next lines to get the outputs of $\mathcal{L}_{CE}$ and $\mathcal{L}^{CNT}_{KD}$.
Then the linear combination of line 14 combines these two losses to get the continuation loss $\mathcal{L}$. Finally, $\mathcal{L}$  is fed into $\Call{Optimization-Back-Propagation}{.}$ function to optimize the student network based on the back propagation of the gradient of this loss function and update the weights of the student. This part of the training is identical to regular training of the neural networks. $\Call{Save-Best-Checkpoint}{.}$ function checks the performance of the current student model on a validation dataset. If it is better than the previous checkpoints, it saves the checkpoint. In the end, we load the best checkpoint and return it.

In the next section, we report the experimental results of Continuation-KD method.
                                                                                         
\begin{algorithm*}[ht]
    \caption{}\label{alg}
    \begin{algorithmic}[1]
        \Function{Continuation-KD}{$S$,$T$,$D$, $\mathcal{T}_{max}$, $n$, $\psi(\cdot)$, $m$}    
          %\Comment{$S$ is student, $T$ is teacher, $X$ is input dataset, $k$ is \#training epochs in each temperature, $\mathcal{T}_{max}$ maximum temperature, $n$ #epochs in stage II}
            % \State $n \gets k \times \mathcal{T}_{max}$ \Comment{n is \#training
            % epochs}
           \State $\mathcal{T} = \mathcal{T}_{max}$ 
           \State $k = \lfloor \frac{n}{\mathcal{T}_{max}}\rfloor$ 
           \For{$i = 1$ \texttt{to $n$}}
               
                 \If{$i \text{ mod } k = 0$}
                    \State $\mathcal{T} = \mathcal{T} - 1$
                \EndIf
                 \State $\Phi \gets 1-\frac{\mathcal{T}-1}{\mathcal{T}_{\text{max}}}$,$\Psi \gets \psi(i)$
                %  \State $\Psi \gets \psi(i)$
                 \State $X, Y \gets$ \Call{Get-Mini-Batch}{D}
                 \State $\mathcal{L}_{CE} \gets CE(\sigma(S(X)), Y)$
                %  \State $\mathcal{L}_{MSE} = \lVert S(X) - \Phi  T(X)\rVert^{2}_{2}$
                 \State $\mathcal{L}^{CNT}_{KD} \gets \max\{0,\lVert S(X) - \Phi  T(X)\rVert^{2}_{2} - m \text{ }\Phi\}$
                 \State $\mathcal{L} = \Psi\mathcal{L}_{CE} + (1 - \Psi)\mathcal{L}^{CNT}_{KD}$
                 \State \Call{Optimization-Back-propagation}{$\mathcal{L}$}
                 \State \Call{Save-Best-Checkpoint}{$S$}
           \EndFor 
            \State $S \gets$ \Call{Load-Best-Checkpoint}{ }

            \State \Return $S$
        \EndFunction
    \end{algorithmic}
\end{algorithm*}

% \begin{algorithm}
% \begin{algorithmic}
% \State $i \gets 10$
% \If{$i\geq 5$} 
%     \State $i \gets i-1$
% \Else
%     \If{$i\leq 3$}
%         \State $i \gets i+2$
%     \EndIf
% \EndIf 
% \end{algorithmic}
% \end{algorithm}

\section{Experiments}
This section demonstrates our evaluation results comparing Continuation-KD with other baselines for natural language processing and computer vision tasks. We compare our method with state-of-the-art techniques such as annealing-KD ~\cite{jafari2021annealing},  TAKD ~\cite{mirzadeh2019improved} and other baselines like Vanilla-KD ~\cite{hinton2015distilling} and training the student only with hard labels. In the following sub-sections, we will discuss each in more detail.

\subsection{Hardware Details}
We trained all our baselines using
a single NVIDIA V100 GPU. All experiments
were run using the PyTorch framework\footnote{https://pytorch.org} and for NLP experiments we used HuggingFace\footnote{https://huggingface.co} API.

\subsection{Image Classification}
For the image classification tasks, we used CIFAR-10 and CIFAR-100 datasets with 10 and 100 classes respectively. Both of these datasets have 60,000 data samples, and each of them is a $32 \times 32$ pixel color image. Also, both of these datasets have 50,000 train and 10,000 test samples. 

% In all of our CV experiments,  ResNet-8 and resNet-110 models are used as the student and the teacher models. Similar experimental setups as the  TAKD method (Mirzadeh et al., 2019) are used for CV experiments. Also, for the TAKD baseline, the ResNet-20 model is used as the TA model. Tables \ref{table:CIFAR_10} and \ref{table:CIFAR_100} depicts the results of these experiments. For all baselines, we trained a ResNet-110 teacher from scratch on the given datasets First. Then we used this teacher in training of all KD baselines. Also, for TAKD, we trained the TA model first by using Vanilla-KD to train a resNet-20 as TA model and then this TA is used to train the resNet-8 student model. Continuation-KD trained for 200 epochs with maximum temprature 20, learning rate 0.2, and batch size 32. Also, the following function used in these experiments:
In all of our computer vision (CV) experiments,  ResNet-8 and ResNet-110 models are used as the student and the teacher respectively. For the TAKD baseline, the ResNet-20 model is used as the TA model. Continuation-KD is trained for 200 epochs with a maximum temperature of 20, a learning rate of 0.2, and batch size 32. Also, the following $\psi(\cdot)$ function used in these experiments:

\begin{equation}
    \begin{split}
        &   \psi(i) =    \begin{cases}
    \frac{i}{150},              &   1 \leq i \leq 150\\
    1,              & i  \geq 150
\end{cases}  
    \end{split}
    \label{eq:phi_prime_cv}
\end{equation}

 The CIFAR-10 and CIFAR-100 experiments results are reported in Tables \ref{table:CIFAR_10} and \ref{table:CIFAR_100}. We can see that Continuation-KD clearly outperforms other baselines. Also, Annealing-KD achieved second-best results in these experiments. However, other baselines showed almost similar performances.

 \begin{table}[ht]
\caption{Comparing the test accuracy of Continuation-KD, Annealing-KD, TAKD, Vanilla-KD, and finetuning on CIFAR-10 dataset with ResNet model} % title of Table
\centering % used for centering table
\begin{tabular}{c c c} % centered columns (4 columns)
\hline
Type &Training method & Accuracy\\
%heading
\hline % inserts single horizontal line
Teacher(110) & from scratch & 93.8\\ % inserting body of the table
TA(20) & KD & 92.39\\
\hline
Student(8) & from scratch & 88.44\\
Student(8) &  KD & 88.45\\
Student(8) & TAKD & 88.47\\
Student(8) & Annealing-KD & 89.44\\
student(8) & \textbf{Continuation-KD} & \textbf{90.21} \\
\hline %inserts single line
\end{tabular}
\label{table:CIFAR_10} % is used to refer this table in the text
\end{table}

\begin{table}[ht]
\caption{Comparing the test accuracy of Continuation-KD, Annealing-KD, TAKD, Vanilla-KD, and finetuning on CIFAR-100 dataset with ResNet model} % title of Table
\centering % used for centering table
\begin{tabular}{c c c} % centered columns (4 columns)
\hline
Type &Training method & Accuracy\\
%heading
\hline % inserts single horizontal line
teacher(110) & from scratch & 71.92\\ % inserting body of the table
TA(20) &  KD & 67.6\\
\hline
student(8) & from scratch & 61.37\\
student(8) &  KD & 61.41\\
student(8) & TAKD & 61.82\\
student(8) & Annealing-KD & 63.1\\
student(8) & \textbf{Continuation-KD} & \textbf{64.2} \\ % is\\ \textbf{64.3} \\

\hline %inserts single line
\end{tabular}
\label{table:CIFAR_100} % is used to refer this table in the text
\end{table}

 \begin{figure*}[bt]
 \centering
  \subfloat[]{\label{figur1:1}\includegraphics[width=0.5\textwidth]{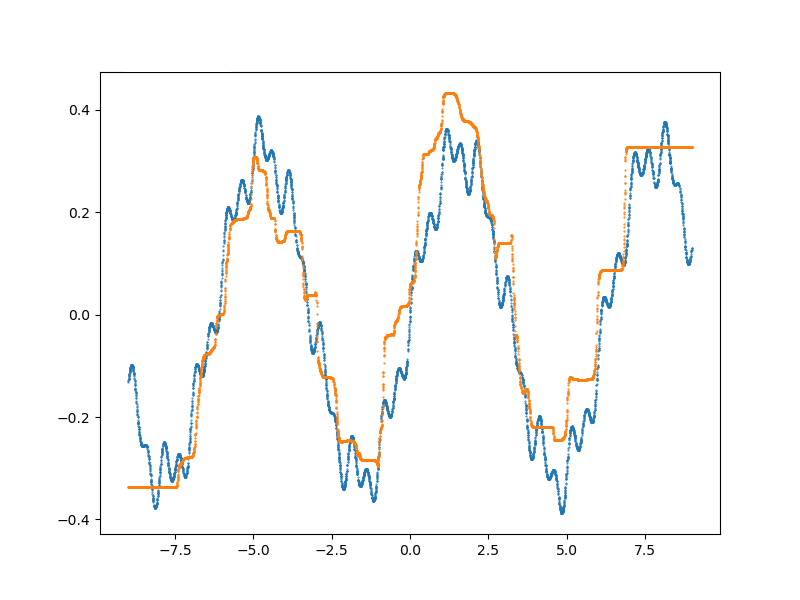}}
  \subfloat[]{\label{figur1:2}\includegraphics[width=0.5\textwidth]{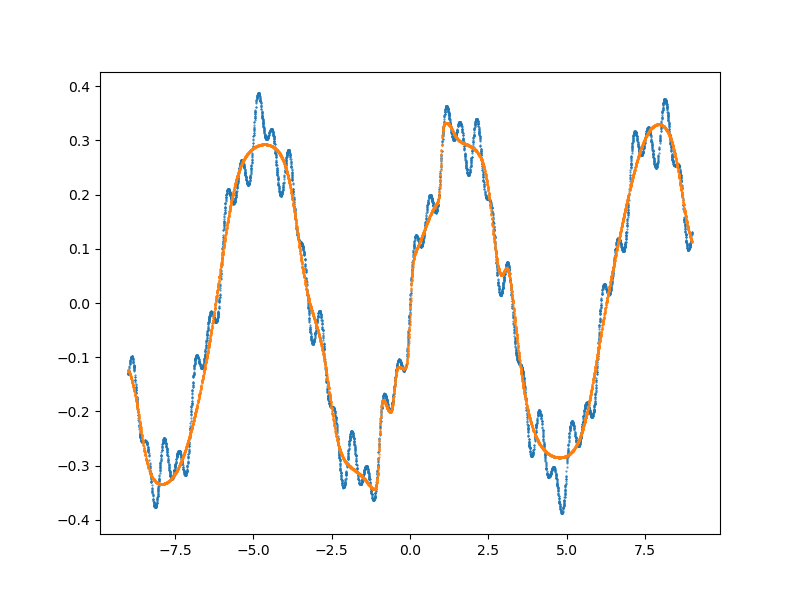}}\\
 
  \caption{(a) Illustrates the behaviour of the student model after training with a noisy teacher without using continuation-KD. (b) Illustrates the behaviour of the student model after training with a noisy teacher by using continuation-KD for training. Blue points are the samples from the noisy teacher. Orange points are the samples of the trained student in each scenario.}
  \label{fig2}
\end{figure*}

 \subsection{Natural Language Understanding}
 \label{NLU_section}
 For the NLU experiments, we use the General Language Understanding Evaluation (GLUE) benchmark \cite{wang2018glue}.  The benchmark contains several tasks, including textual entailment (RTE and MNLI), question-answer entailment (QNLI), paraphrase (MRPC), question paraphrase (QQP), sentiment (SST-2), textual similarity (STS-B), linguistic acceptability (CoLA), and Winograd Schema (WNLI).
 
 Two types of students with different capacities are used in these experiments. In our first experiment, we used distilRoBERTa with 6 layers as our student and RoBERTa-Large (24 layers) as our teacher. Also, for the TAKD baseline, we used RoBERTa-base (12 layers) as the teacher assistant model. For this experiment, the maximum temperature was  10,  learning rate was 2e-5, and batch size was 64. Also, we used the pre-trained distilRoBERTa as our student and we fine-tuned it for 30 epochs. The $\psi$ function for this experiment is defined as following:
 \begin{equation}
    \begin{split}
        &   \psi(i) =    \begin{cases}
    \frac{i}{40},              &   1 \leq i \leq 20\\
    1,              & 20 < i  \leq 30
\end{cases}  
    \end{split}
    \label{eq:phi_prime_nlp}
\end{equation}
 
 Tables \ref{table:roBERTa_dev} and \ref{table:roBERTa_test} show the results of this experiment on the dev set and test set. As shown in these tables, Continuation-KD achieved superior results in most of the tasks and it improves the previous baselines with a good overall average score.

For the second experiment, we utilize BERT-large (24-layers) as the teacher, BERT-Small (4-layers) as the student and BERT-base (12-layers) as the teacher-assistant for TAKD. We train the teacher for 30 epochs, and for the student, we use the same hyperparameters as the first experiment. We report the results in Tables \ref{table:small-bert-dev} and \ref{table:small-bert-test}. Here, Continuation-KD also outperforms other baselines and gets better overall performance than its competitors.

All presented results in our experiments show that continuation-KD performs better than TAKD and Annealing-KD, which indicates the effectiveness of this method. The experimental results support the claim that our proposed Continuation-KD can provide a better generalization than the other methods.
 
\begin{table*}[hbt!]
\caption{DistilRoBERTa results for Continuation-KD on dev set. F1 scores are reported for MRPC, pearson correlations for STB-B, and accuracy scores for all other tasks.} % title of Table
\centering % used for centering table
\begin{tabular}{c c c c c c c c c c} % centered columns (4 columns)
\hline
KD Method & CoLA & RTE & MRPC & STS-B & SST-2 & QNLI & QQP & MNLI & score \\ [0.5ex] % inserts table
%heading
\hline % inserts single horizontal line
Teacher & 68.1 & 86.3 & 91.9 & 92.3 & 96.4 & 94.6 & 91.5 & 90.22 & 88.91 \\
\hline
Finetune & 59.3 & 67.9 & 88.6 & 88.5 & 92.5 & 90.8 & 90.9 & 84 & 82.81 \\
Vanilla-KD & 60.97 & 71.11 & 90.2 & 88.86 & 92.54 & 91.37 & 91.64 & 84.18 & 83.85 \\
TAKD & 61.15 & 71.84 & 89.91 & 88.94 & 92.54 & 91.32 & \textbf{91.7} & 83.89 & 83.91 \\
Annealing KD & 61.67 & 73.64 & 90.6 & 89.01 & 93.11 & 91.64 & 91.5 & \textbf{85.34} & 84.56 \\
Continuation-KD & \textbf{63.75} & \textbf{77.62} & \textbf{91.93} & \textbf{89.71} & \textbf{93.58} & \textbf{91.76} & 91.64 & 85.16 & \textbf{85.64} \\ 

\hline %inserts single line
\end{tabular}
\label{table:roBERTa_dev} % is used to refer this table in the text
\end{table*}

\begin{table*}[hbt!]
\caption{Performance of DistilRoBERTa trained by Continuation-KD on the GLUE leaderboard compared with Vanilla-KD, TAKD, and annealing KD.} % title of Table
\centering % used for centering table
\begin{tabular}{c c c c c c c c c c} % centered columns (4 columns)
\hline
KD Method & CoLA & RTE & MRPC & STS-B & SST-2 & QNLI & QQP & MNLI & score \\ [0.5ex] % inserts table
%heading
\hline % inserts single horizontal line
Finetune & 52.97 & 74.56 & 87.93 & 84.9 & 93.1 & 90.57 & 83.4 & 88.77 & 82.02 \\
Vanilla-KD & 54.3 & 74.1 & 86 & 85.7 & 93.1 & 83.6 & 90.8 & 89.5 & 82.14 \\
TAKD & 53.2 & 74.2 & 86.7 & 85.6 & 93.2 & 83.8 & 91 & 89.4 & 82.14 \\
Annealing KD & 54 & 73.7 & 88 & \textbf{87.0} & 93.6 & 83.8 & 90.8 & 89.7 & 82.58 \\
Continuation-KD & \textbf{54.5} & 74.2 & \textbf{90} & \textbf{87.0} & \textbf{93.8} & \textbf{84.7} & \textbf{91.6} & \textbf{90.1} & \textbf{83.24} \\

\hline %inserts single line
\end{tabular}
\label{table:roBERTa_test} % is used to refer this table in the text
\end{table*}

\begin{table*}[hbt!]
\caption{BERT-Small results for Continuation-KD on dev set. F1 scores are reported for MRPC, pearson correlations
for STS-B, and accuracy scores for all other tasks.} % title of Table
\centering % used for centering table
\begin{tabular}{c c c c c c c c c c} % centered columns (4 columns)
\hline
KD Method & CoLA & RTE & MRPC & STS-B & SST-2 & QNLI & QQP & MNLI & score \\
%heading
\hline % inserts single horizontal line
Teacher & 65.8 & 71.48 & 91.38 & 89.2 & 92.77 & 92.82 & 91.45 & 86.3 & 85.15\\
\hline
Finetune & 41.7 & 64.98 & 83.75 & 87.41 & 88.3 & 86.49 & 88.43 & 78.42 & 77.44\\
Vanilla-KD & \textbf{41.89} & 64.98 & 86 & 85.95 & 88.76 & 86.75 & 88.24 & 78.62 & 77.65\\
TAKD & 40.2 & 65.7 & 85.23 & 86.44 & 88.88 & 86.78 & 88.4 & 78.78 & 77.55\\
Annealing-KD & 41.36 & 64.9 & 87.93 & 87.04 & 89.56 & 86.99 & 88.58 & 78.66 & 78.13\\
Continuation-KD & 41.51 & \textbf{65.26} & \textbf{89.44} & \textbf{87.81} & \textbf{90.21} & \textbf{87.52} & \textbf{90.61} & \textbf{79.27} & \textbf{78.95}\\
\hline %inserts single line
\end{tabular}
\label{table:small-bert-dev} % is used to refer this table in the text
\end{table*}

\begin{table*}[hbt!]
\caption{BERT-Small results for Continuation-KD on test set. F1 scores are reported for MRPC, pearson correlations
for STS-B, and accuracy scores for all other tasks.} % title of Table
\centering % used for centering table
\begin{tabular}{c c c c c c c c c c} % centered columns (4 columns)
\hline
KD Method & CoLA & RTE & MRPC & STS-B & SST-2 & QNLI & QQP & MNLI & score \\
%heading
\hline % inserts single horizontal line
Finetune & 38.1 & 61.8 & 83.4 & 78.8 & 89.7 & 86.4 & 78.1 & 77.6 & 74.24\\
Vanilla-kd & 37.3 & 63.4 & 80.6 & 78.2 & 90.2 & 86.5 & 78.3 & 78.3 & 74.09\\
TAKD & 38.5 & 62.3 & 80.5 & 79.3 & 89.7 & 86.7 & 78 & 78.2 & 74.14\\
Annealing-KD & 38.3 & 63.3 & 81.9 & 80.6 & 89.8 & 86.8 & 78.4 & 79.3 & 74.78\\
Continuation-KD & 38.5 & \textbf{64.8} & \textbf{84.6} & \textbf{82.5} & \textbf{90.6} & \textbf{87.2} & \textbf{80.1} & \textbf{79.5} & \textbf{75.98}\\
\hline %inserts single line
\end{tabular}
\label{table:small-bert-test} % is used to refer this table in the text
\end{table*}

% \begin{table}[ht]
% \caption{Comparing the test accuracy of Continuation-KD, Annealing-KD, TAKD, Vanilla-KD, and finetuning on CIFAR-10 dataset with ResNet model } % title of Table
% \centering % used for centering table
% \resizebox{\columnwidth}{!}{%
% \begin{tabular}{c c c c} % centered columns (4 columns)
% \Xhline{4\arrayrulewidth} %inserts double horizontal lines
% Model & Type & Training method & Accuracy\\ [0.5ex] % inserts table
% %heading
% \hline % inserts single horizontal line
% Teacher(110) & from scratch & 93.8\\ % inserting body of the table
% & TA(20) & KD & 92.39\\
% \cline{2-4}
% & Student(8) & from scratch & 88.44\\
% & Student(8) &  KD & 88.45\\
% & Student(8) & TAKD & 88.47\\
% & Student(8) & \textbf{Annealing KD (ours)} & \textbf{89.44}\\
% \hline %inserts single line
% \multirow{6}{*}{CNN} & Teacher(10) & from scratch & 90.1\\ % inserting body of the table
%  & TA(4) &  KD & 82.39\\
%  \cline{2-4}
%  & Student(2) & from scratch & 72.75\\
%   & Student(2) &  KD & 72.43\\
%   & Student(2) & TAKD & 72.62\\
%   & Student(2) & \textbf{Annealing KD (ours)} & \textbf{73.17}\\
% \hline %inserts single line
% \end{tabular}
% }
% \label{tableCV1:nonlin} % is used to refer this table in the text
% \end{table}

\section{Analysis}
To investigate how Continuation-KD works, we did two ablation studies explaining different aspects of this technique: effects of the dynamically changing factors and noise mitigation.

\paragraph{Effects of Dynamic Factors $\phi$ and $\psi$.}
In the first ablation, we scrutinize the effect of each dynamically changing component of the Continuation-KD loss function. It basically considers the effect of functions:  $\psi(i)$ from Eq.~\ref{eq:Continuation},  $\phi(\mathcal{T}_i)$  when it anneals the outputs of the teacher in Eq.~\ref{eq:Continuation_loss}, and  $\phi(\mathcal{T}_i)$  when it anneals the margin of the hinge loss in Eq.~\ref{eq:Continuation_loss}. To investigate the effects of these components, we repeat the NLU experiments with distilRoBERTa model described in section~\ref{NLU_section} on MRPC and RTE datasets three times. In each trial, we fixed two of the components, and we only let one of them dynamically change during training.  
Table \ref{table:abalation} reports the performance of each of these experiments. The first three columns of this table show the performance of distilRoBERTa on each of the datasets when only one of the three dynamical components changes and two others are fixed. The last column shows the performance of the model on these datasets in a regular training when all components dynamically change. In the first column,  the $\phi = 1$ is fixed for both teacher and margin. In the second column, the margin's coefficient is fixed to 1 and $\psi = 0.5$. Also, in the third column, the teacher's coefficient is fixed to 1 and $\psi = 0.5$. 

As shown in Table \ref{table:abalation}, for both datasets, the performance in the first three columns is almost similar, which indicates the equal contribution of each dynamic component in the improvement of the results. Also, the last column shows the dramatic improvement in the performance. Hence, we can conclude from this experiment that all three dynamic components are necessary for achieving better performance.

\begin{table}[ht]
\caption{Performance of distilRoBERTa on MRPC and RTE datasets for different dynamically changing components of continuation-KD loss} % title of Table
\centering % used for centering table
\resizebox{\columnwidth}{!}{
\begin{tabular}{c c c c c} % centered columns (4 columns)
\hline
dataset &$\psi$ & $\phi$ (teacher) & $\phi$ (margin) & all\\
%heading
\hline % inserts single horizontal line
MRPC & 91.09 & 91.21 & 90.90 & 91.93\\ 
RTE  & 72.56 & 73.64 & 71.48 & 77.62\\ % inserting body of the table
\hline %inserts single line
\end{tabular}
}
\label{table:abalation} % is used to refer this table in the text
\end{table}

\paragraph{Noise Robustness} In the second ablation study, we visualize the effect of Continuation-KD with a noisy teacher. For this purpose, we consider a sinusoidal function with low frequency, and then we add to it a sinusoidal noise with high frequency (blue curves in Figure~\ref{fig2}). Then, we take a fully connected network with one-dimensional input and output and two hidden layers with 128 neurons in each layer as a  student model. We sample 3,000 points from the graph of the noisy sinusoidal function and train the student model once with Vanilla-KD and once with Continuation-KD (orange curves in Figure~\ref{fig2}). As Figure~\ref{fig2} shows, the model trained with Continuation-KD has a much smoother curve in comparison with the model trained with vanilla-KD and could learn the main behaviour of the teacher function rather than learning noise.  

% The first ablation study indicates that all of the dynamical components designed in continuation-KD are important and have an effect on improving the results during training.  Also, the second investigation supports our initial claim about the effectiveness of continuation-KD in avoiding noise modelling and helping the student model to learn the main behaviour of the teacher function.

\section{Conclusion}
In this work, we present Continuation-KD, a novel KD method inspired by Continuation optimization. Our Continuation-KD technique starts optimizing a smoothed version of the objective function and gradually increases the complexity of the loss towards the original, highly non-convex one. We demonstrated that our method alleviates the capacity gap problem: an innate problem of KD resulting from the different capacities of a student's and a teacher's networks which detriments the performance. Besides that, we show that the method can lessen the student's overfitting to noise in the teacher's output. Our technique is stable because it doesn't require two stages in the training and is efficient. It outperforms its competitor KD methods for different backbone models' architectures in both computer vision and NLP. 

For this investigation, we proposed to implement Continuation method by smoothing the objective with the hinge loss. However, it can potentially be done differently. Investigating other realizations of Continuation Optimization for improving small models' performance is an interesting next step.

\section*{Limitation}
One of the advantages of our proposed method is that it mitigates the noise in the teacher's output and prevents the student from overfitting to the noise. We empirically demonstrate this claim, but we don't have rigorous theoretical proof of how continuation method achieves this robustness.

\section*{Acknowledgments}
We thank Mindspore\footnote{A new deep learning computing framework \url{https://www.mindspore.cn/}} for the partial support of this work. 
We thank the anonymous reviewers for their insightful comments.

\bibliography{anthology,custom}

\begin{thebibliography}{31}
\expandafter\ifx\csname natexlab\endcsname\relax\def\natexlab#1{#1}\fi

\bibitem[{Bie et~al.(2019)Bie, Venkitesh, Monteiro, Haidar, Rezagholizadeh
  et~al.}]{bie2019simplified}
Alex Bie, Bharat Venkitesh, Joao Monteiro, Md~Haidar, Mehdi Rezagholizadeh,
  et~al. 2019.
\newblock A simplified fully quantized transformer for end-to-end speech
  recognition.
\newblock \emph{arXiv preprint arXiv:1911.03604}.

\bibitem[{Brown et~al.(2020{\natexlab{a}})Brown, Mann, Ryder, Subbiah, Kaplan,
  Dhariwal, Neelakantan, Shyam, Sastry, Askell, Agarwal, Herbert-Voss, Krueger,
  Henighan, Child, Ramesh, Ziegler, Wu, Winter, Hesse, Chen, Sigler, Litwin,
  Gray, Chess, Clark, Berner, McCandlish, Radford, Sutskever, and
  Amodei}]{NEURIPS2020_1457c0d6}
Tom Brown, Benjamin Mann, Nick Ryder, Melanie Subbiah, Jared~D Kaplan, Prafulla
  Dhariwal, Arvind Neelakantan, Pranav Shyam, Girish Sastry, Amanda Askell,
  Sandhini Agarwal, Ariel Herbert-Voss, Gretchen Krueger, Tom Henighan, Rewon
  Child, Aditya Ramesh, Daniel Ziegler, Jeffrey Wu, Clemens Winter, Chris
  Hesse, Mark Chen, Eric Sigler, Mateusz Litwin, Scott Gray, Benjamin Chess,
  Jack Clark, Christopher Berner, Sam McCandlish, Alec Radford, Ilya Sutskever,
  and Dario Amodei. 2020{\natexlab{a}}.
\newblock \href
  {https://proceedings.neurips.cc/paper/2020/file/1457c0d6bfcb4967418bfb8ac142f64a-Paper.pdf}
  {Language models are few-shot learners}.
\newblock In \emph{Advances in Neural Information Processing Systems},
  volume~33, pages 1877--1901. Curran Associates, Inc.

\bibitem[{Brown et~al.(2020{\natexlab{b}})Brown, Mann, Ryder, Subbiah, Kaplan,
  Dhariwal, Neelakantan, Shyam, Sastry, Askell et~al.}]{brown2020language}
Tom~B Brown, Benjamin Mann, Nick Ryder, Melanie Subbiah, Jared Kaplan, Prafulla
  Dhariwal, Arvind Neelakantan, Pranav Shyam, Girish Sastry, Amanda Askell,
  et~al. 2020{\natexlab{b}}.
\newblock Language models are few-shot learners.
\newblock \emph{arXiv preprint arXiv:2005.14165}.

\bibitem[{Chebotar and Waters(2016)}]{chebotar2016distilling}
Yevgen Chebotar and Austin Waters. 2016.
\newblock Distilling knowledge from ensembles of neural networks for speech
  recognition.
\newblock In \emph{Interspeech}, pages 3439--3443.

\bibitem[{Clark et~al.(2019)Clark, Luong, Khandelwal, Manning, and
  Le}]{clark-etal-2019-bam}
Kevin Clark, Minh-Thang Luong, Urvashi Khandelwal, Christopher~D. Manning, and
  Quoc~V. Le. 2019.
\newblock \href {https://doi.org/10.18653/v1/P19-1595} {{BAM}! born-again
  multi-task networks for natural language understanding}.
\newblock In \emph{Proceedings of the 57th Annual Meeting of the Association
  for Computational Linguistics}, pages 5931--5937, Florence, Italy.
  Association for Computational Linguistics.

\bibitem[{Davidenko(1953)}]{continuation_53}
D.F. Davidenko. 1953.
\newblock On a new method of numerical solution of systems of nonlinear
  equations.
\newblock \emph{Dokl. Akad. Nauk SSSR}, 88:601--602.

\bibitem[{Devlin et~al.(2018)Devlin, Chang, Lee, and
  Toutanova}]{devlin2018bert}
Jacob Devlin, Ming-Wei Chang, Kenton Lee, and Kristina Toutanova. 2018.
\newblock Bert: Pre-training of deep bidirectional transformers for language
  understanding.
\newblock \emph{arXiv preprint arXiv:1810.04805}.

\bibitem[{Dosovitskiy et~al.(2020)Dosovitskiy, Beyer, Kolesnikov, Weissenborn,
  Zhai, Unterthiner, Dehghani, Minderer, Heigold, Gelly
  et~al.}]{dosovitskiy2020image}
Alexey Dosovitskiy, Lucas Beyer, Alexander Kolesnikov, Dirk Weissenborn,
  Xiaohua Zhai, Thomas Unterthiner, Mostafa Dehghani, Matthias Minderer, Georg
  Heigold, Sylvain Gelly, et~al. 2020.
\newblock An image is worth 16x16 words: Transformers for image recognition at
  scale.
\newblock \emph{arXiv preprint arXiv:2010.11929}.

\bibitem[{Gulcehre et~al.(2017)Gulcehre, Moczulski, Visin, and
  Bengio}]{mollifying}
Caglar Gulcehre, Marcin Moczulski, Francesco Visin, and Yoshua Bengio. 2017.
\newblock \href {https://openreview.net/forum?id=r1G4z8cge} {Mollifying
  networks}.
\newblock In \emph{5th International Conference on Learning Representations}.

\bibitem[{Guo et~al.(2020)Guo, Wang, Wu, Yu, Liang, Hu, and
  Luo}]{guo2020online}
Qiushan Guo, Xinjiang Wang, Yichao Wu, Zhipeng Yu, Ding Liang, Xiaolin Hu, and
  Ping Luo. 2020.
\newblock Online knowledge distillation via collaborative learning.
\newblock In \emph{Proceedings of the IEEE/CVF Conference on Computer Vision
  and Pattern Recognition}, pages 11020--11029.

\bibitem[{Haidar et~al.(2021)Haidar, Anchuri, Rezagholizadeh, Ghaddar,
  Langlais, and Poupart}]{haidar2021rail}
Md~Akmal Haidar, Nithin Anchuri, Mehdi Rezagholizadeh, Abbas Ghaddar, Philippe
  Langlais, and Pascal Poupart. 2021.
\newblock Rail-kd: Random intermediate layer mapping for knowledge
  distillation.
\newblock \emph{arXiv preprint arXiv:2109.10164}.

\bibitem[{Hinton et~al.(2015)Hinton, Vinyals, and Dean}]{hinton2015distilling}
Geoffrey Hinton, Oriol Vinyals, and Jeff Dean. 2015.
\newblock Distilling the knowledge in a neural network.
\newblock \emph{arXiv preprint arXiv:1503.02531}.

\bibitem[{Jacob et~al.(2018)Jacob, Kligys, Chen, Zhu, Tang, Howard, Adam, and
  Kalenichenko}]{jacob2018quantization}
Benoit Jacob, Skirmantas Kligys, Bo~Chen, Menglong Zhu, Matthew Tang, Andrew
  Howard, Hartwig Adam, and Dmitry Kalenichenko. 2018.
\newblock Quantization and training of neural networks for efficient
  integer-arithmetic-only inference.
\newblock In \emph{Proceedings of the IEEE Conference on Computer Vision and
  Pattern Recognition}, pages 2704--2713.

\bibitem[{Jafari et~al.(2021)Jafari, Rezagholizadeh, Sharma, and
  Ghodsi}]{jafari2021annealing}
Aref Jafari, Mehdi Rezagholizadeh, Pranav Sharma, and Ali Ghodsi. 2021.
\newblock Annealing knowledge distillation.
\newblock In \emph{Proceedings of the 16th Conference of the European Chapter
  of the Association for Computational Linguistics: Main Volume}, pages
  2493--2504.

\bibitem[{Jiao et~al.(2019)Jiao, Yin, Shang, Jiang, Chen, Li, Wang, and
  Liu}]{jiao2019tinybert}
Xiaoqi Jiao, Yichun Yin, Lifeng Shang, Xin Jiang, Xiao Chen, Linlin Li, Fang
  Wang, and Qun Liu. 2019.
\newblock Tinybert: Distilling bert for natural language understanding.
\newblock \emph{arXiv preprint arXiv:1909.10351}.

\bibitem[{Kamalloo et~al.(2021)Kamalloo, Rezagholizadeh, Passban, and
  Ghodsi}]{kamalloo2021not}
Ehsan Kamalloo, Mehdi Rezagholizadeh, Peyman Passban, and Ali Ghodsi. 2021.
\newblock Not far away, not so close: Sample efficient nearest neighbour data
  augmentation via minimax.
\newblock \emph{arXiv preprint arXiv:2105.13608}.

\bibitem[{Liu et~al.(2019)Liu, Ott, Goyal, Du, Joshi, Chen, Levy, Lewis,
  Zettlemoyer, and Stoyanov}]{liu2019roberta}
Yinhan Liu, Myle Ott, Naman Goyal, Jingfei Du, Mandar Joshi, Danqi Chen, Omer
  Levy, Mike Lewis, Luke Zettlemoyer, and Veselin Stoyanov. 2019.
\newblock Roberta: A robustly optimized bert pretraining approach.
\newblock \emph{arXiv preprint arXiv:1907.11692}.

\bibitem[{Lopez-Paz et~al.(2015)Lopez-Paz, Bottou, Sch{\"o}lkopf, and
  Vapnik}]{lopez2015unifying}
David Lopez-Paz, L{\'e}on Bottou, Bernhard Sch{\"o}lkopf, and Vladimir Vapnik.
  2015.
\newblock Unifying distillation and privileged information.
\newblock \emph{arXiv preprint arXiv:1511.03643}.

\bibitem[{Mirzadeh et~al.(2019)Mirzadeh, Farajtabar, Li, and
  Ghasemzadeh}]{mirzadeh2019improved}
Seyed-Iman Mirzadeh, Mehrdad Farajtabar, Ang Li, and Hassan Ghasemzadeh. 2019.
\newblock Improved knowledge distillation via teacher assistant: Bridging the
  gap between student and teacher.
\newblock \emph{arXiv preprint arXiv:1902.03393}.

\bibitem[{Mobahi and Fisher~III(2015)}]{mobahi_fisher_2015}
Hossein Mobahi and John Fisher~III. 2015.
\newblock \href {https://doi.org/10.1609/aaai.v29i1.9356} {A theoretical
  analysis of optimization by gaussian continuation}.
\newblock \emph{Proceedings of the AAAI Conference on Artificial Intelligence},
  29(1).

\bibitem[{Rashid et~al.(2020)Rashid, Lioutas, Ghaddar, and
  Rezagholizadeh}]{rashid2020towards}
Ahmad Rashid, Vasileios Lioutas, Abbas Ghaddar, and Mehdi Rezagholizadeh. 2020.
\newblock Towards zero-shot knowledge distillation for natural language
  processing.
\newblock \emph{arXiv preprint arXiv:2012.15495}.

\bibitem[{Rashid et~al.(2021)Rashid, Lioutas, and
  Rezagholizadeh}]{rashid2021mate}
Ahmad Rashid, Vasileios Lioutas, and Mehdi Rezagholizadeh. 2021.
\newblock Mate-kd: Masked adversarial text, a companion to knowledge
  distillation.
\newblock \emph{arXiv preprint arXiv:2105.05912}.

\bibitem[{Sanh et~al.(2019)Sanh, Debut, Chaumond, and
  Wolf}]{sanh2019distilbert}
Victor Sanh, Lysandre Debut, Julien Chaumond, and Thomas Wolf. 2019.
\newblock Distilbert, a distilled version of bert: smaller, faster, cheaper and
  lighter.
\newblock \emph{arXiv preprint arXiv:1910.01108}.

\bibitem[{Sun et~al.()Sun, Yu, Song, Liu, Yang, and Zhou}]{sunmobilebert}
Zhiqing Sun, Hongkun Yu, Xiaodan Song, Renjie Liu, Yiming Yang, and Denny Zhou.
\newblock Mobilebert: Task-agnostic compression of bert for resource limited
  devices.

\bibitem[{Tjandra et~al.(2018)Tjandra, Sakti, and Nakamura}]{tjandra2018tensor}
Andros Tjandra, Sakriani Sakti, and Satoshi Nakamura. 2018.
\newblock Tensor decomposition for compressing recurrent neural network.
\newblock In \emph{2018 International Joint Conference on Neural Networks
  (IJCNN)}, pages 1--8. IEEE.

\bibitem[{Turc et~al.(2019)Turc, Chang, Lee, and Toutanova}]{turc2019wellread}
Iulia Turc, Ming-Wei Chang, Kenton Lee, and Kristina Toutanova. 2019.
\newblock Well-read students learn better: On the importance of pre-training
  compact models.
\newblock \emph{arXiv preprint arXiv:1908.08962}.

\bibitem[{Vaswani et~al.(2017)Vaswani, Shazeer, Parmar, Uszkoreit, Jones,
  Gomez, Kaiser, and Polosukhin}]{vaswani2017attention}
Ashish Vaswani, Noam Shazeer, Niki Parmar, Jakob Uszkoreit, Llion Jones,
  Aidan~N Gomez, {\L}ukasz Kaiser, and Illia Polosukhin. 2017.
\newblock Attention is all you need.
\newblock In \emph{Advances in neural information processing systems}, pages
  5998--6008.

\bibitem[{Wang et~al.(2018)Wang, Singh, Michael, Hill, Levy, and
  Bowman}]{wang2018glue}
Alex Wang, Amanpreet Singh, Julian Michael, Felix Hill, Omer Levy, and Samuel~R
  Bowman. 2018.
\newblock Glue: A multi-task benchmark and analysis platform for natural
  language understanding.
\newblock \emph{arXiv preprint arXiv:1804.07461}.

\bibitem[{Watson(2000)}]{continuation_00}
L.T. Watson. 2000.
\newblock Theory of globally convergent probability-one homotopies for
  nonlinear programming.
\newblock \emph{SIAM Journal on Optimization}, 11(3):761--780.

\bibitem[{Wu et~al.(2021)Wu, Rezagholizadeh, Ghaddar, Haidar, and
  Ghodsi}]{wu2021universal}
Yimeng Wu, Mehdi Rezagholizadeh, Abbas Ghaddar, Md~Akmal Haidar, and Ali
  Ghodsi. 2021.
\newblock Universal-kd: Attention-based output-grounded intermediate layer
  knowledge distillation.
\newblock In \emph{Proceedings of the 2021 Conference on Empirical Methods in
  Natural Language Processing}, pages 7649--7661.

\bibitem[{Yun et~al.(2020)Yun, Park, Lee, and Shin}]{yun2020regularizing}
Sukmin Yun, Jongjin Park, Kimin Lee, and Jinwoo Shin. 2020.
\newblock Regularizing class-wise predictions via self-knowledge distillation.
\newblock In \emph{Proceedings of the IEEE/CVF conference on computer vision
  and pattern recognition}, pages 13876--13885.

\end{thebibliography}
% \bibliography{custom}
\bibliographystyle{acl_natbib}

\appendix

% \section{Example Appendix}
% \label{sec:appendix}

% This is a section in the appendix.

\end{document}